\documentclass[letterpaper, 10 pt, conference]{ieeeconf}

\IEEEoverridecommandlockouts

\usepackage{times}
\usepackage{amsmath}
\usepackage{amssymb}

\title{\LARGE \bf
Robust Plan Execution with Unexpected Observations
}

\usepackage{color}
\usepackage{algorithm}
\usepackage{algorithmicx}
\usepackage[noend]{algpseudocode}
\usepackage{graphicx}
\usepackage{todonotes}
\usepackage{fancyvrb}
\usepackage[capitalize]{cleveref}
\usepackage{subcaption}
\usepackage{multirow}
\usepackage{array}

\usepackage{amsthm}

\algnewcommand\algorithmicforeach{\textbf{for each}}
\algdef{S}[FOR]{ForEach}[1]{\algorithmicforeach\ #1\ \algorithmicdo}

\newcommand{\eff}{\mathit{eff}}
\newcommand{\prob}{\ensuremath{\rho}}

\newtheorem{definition}{Definition}[section]

\usepackage{tikz}
\usetikzlibrary{positioning, shapes, calc}

\renewcommand{\baselinestretch}{0.99}

\author{Oscar Lima$^{1,2}$, Michael Cashmore$^{3}$, Daniele Magazzeni$^{3}$, Andrea Micheli$^{4}$ and Rodrigo Ventura$^{1}$
\thanks{$^{1}$ Institute for Systems and Robotics, Instituto Superior Tecnico, Lisbon}%
\thanks{$^{2}$ DFKI Robotics Innovation Center, Bremen, Germany}%
\thanks{$^{3}$ King's College London, UK}%
\thanks{$^{4}$ Fondazione Bruno Kessler, Trento, Italy}%
}

\begin{document}

\maketitle
\thispagestyle{empty}
\pagestyle{empty}

\begin{abstract}
In order to ensure the robust actuation of a plan, execution must be adaptable to unexpected situations in the world and to exogenous events. This is critical in domains in which committing to a wrong ordering of actions can cause the plan failure, even when all the actions succeed. We propose an approach to the execution of a task plan that permits some adaptability to unexpected observations of the state while maintaining the validity of the plan through online reasoning.

Our approach computes an adaptable, partially-ordered plan from a given totally-ordered plan. The partially-ordered plan is adaptable in that it can exploit beneficial differences between the world and what was expected. The approach is general in that it can be used with any task planner that produces either a totally or a partially-ordered plan.
We propose a plan execution algorithm that computes online the complete set of valid totally-ordered plans described by an adaptable partially-ordered plan together with the probability of success for each of them. This set is then used to choose the next action to execute.
\end{abstract}

\section{Introduction}

Robust task plan execution is a fundamental problem in the intersection of AI planning and Robotics: the execution of the planned course of actions in the real world may differ from what was expected at planning time. A classical example of this discrepancy is the duration of actions that often depend upon external factors, impossible to model in the planning domain. In addition, plans are usually generated under the assumption of a static world that does not change without performing an action. In reality, however, the world in which the robot operates is often dynamic and comprises exogenous events, i.e. events that are not under the control of the agent and may happen unexpectedly. Such events can interfere with the planned course of actions and therefore we need to monitor the execution by means of observations and possibly adapt the decisions to cope with such contingencies.

Nonetheless, in many situations, minor adjustments to the plan can be sufficient to retain validity with respect to the ground truth and reach the plan objective. For example, several techniques have been devised to absorb small variations in action durations.
This is not the case when the planning problem exhibits temporal deadlines, time-windows or synchronizations, because minor delays could impact the success of the plan. This is particularly relevant when concurrency is considered, e.g. in multi-robot domains, because there can be positive and negative interactions between parallel actions. Hence, ordering constraints that arise from the coordination of these actions must be considered at execution time.

In this paper, we propose a novel flow from AI Planning to action execution aiming at the following research question:
\begin{quote}
Since observations during execution may differ from what was expected at planning time (including action duration and propositional state), is the plan valid and what is the next execution choice that maximizes probability of reaching the goal?
\end{quote}
Our approach (described in \cref{sec:method}) starts from a totally-ordered plan by extracting an adaptable partially-ordered plan as an offline step. Differently from other approaches, we allow some causal constraints to be violated in order to allow for a stronger run-time adaptation. Then, we define an online algorithm that, given an estimation of the probabilities of each planning variable, analyzes all the valid totally-ordered plans induced by the adaptable partially-ordered plan, associating a probability of success to each of them. In turn, this set of totally-ordered plans is used by a novel action selection policy to choose the next action to execute that maximizes the probability of achieving the planning goal during execution. This execution flow is extremely flexible because, depending on the observations, it allows for dynamic re-ordering of the planned set of actions as well as the skipping of actions that might no longer be needed. 

In \cref{sec:exp} we describe how the approach is integrated into the planning and execution framework ROSPlan~\cite{cas15} and we empirically demonstrate in simulation that our approach leads to a consistently fewer re-plannings, and results in fewer actions executed. Moreover, we show how, despite being theoretically demanding in terms of performance, the whole technique can be implemented to be a fast action selection policy for practical use-cases.

\subsection{Related Work}

There has been considerable work in the literature concerning the robust execution of plans.
Some authors proposed ways to increase the flexibility of temporal plans to cope with more situations at runtime~\cite{bac98,frank-linear-envelope,flexibility-do}; others devised techniques to synthesize correct-by-construction flexible plans~\cite{ixtet,europa,apsi,platinum}. In this paper, we relax a fundamental constraint that has been at the base of these previous works: we break the causal structure of the plan by discarding  causal constraints in order to allow for more run-time adaptability. The obtained plan admits executions that are invalid for the planning model, therefore we employ a runtime action selection policy that dynamically selects actions that are causally-valid and are likely to reach the goal. This effectively moves the causal reasoning online instead of limiting the executor to blindly follow the causal structure prescribed by the plan.


The authors of \cite{lev18} show that a Temporal Plan Network Under Uncertainty is an encoding of a set of many different candidate STN~\cite{dec91} sub-plans. They define an \textit{correct execution} as an ordering of activities that is causally complete (each event's conditions are satisfied) and temporally consistent with respect to the STN.
Our approach is strongly related to this idea: we solve the problem of generating the complete set of \textit{valid executions} for an adaptable partially-ordered plan. A valid execution is a \textit{correct execution} which also achieves the goal.

This problem is also described in \cite{kim01} in the form of selecting an execution of a Temporal Plan Network compiled from RMPL, and in \cite{cim16} in the form of synthesizing a dynamically controllable strategy for a disjunctive temporal network with uncertainty. During execution, we tackle a similar problem, without considering decision nodes, and accounting for the selected execution's probability of success. In addition, our approach allows the plan execution to adapt to some unexpected observations.

Uncontrollable temporal durations can be also addressed via strong controllability~\cite{kar15}. Instead, \cite{cas19} proposes a technique that given a plan parameterized with $n$ temporal durations and domain constants automatically generates an $n$-dimensional region corresponding to a valid execution. The latter work generates a region over real-valued parameters for which the plan remains valid. Our work differs because we generate the set of total orderings for which the plan is valid and attach to each of these a probability of success; this copes with discrepancies in the discrete state that are not covered by \cite{cas19}.

\section{Background}\label{sec:met}

We start by formalizing our definition of a planning problem and a plan using the PDDL2.1 formalism~\cite{fox03}.

\begin{definition}
A \textbf{planning problem} is a tuple $\langle P, V, A, I, G \rangle$ where
$P$ is a set of propositions;
$V$ is a set of real variables;
$A$ is a set of durative and instantaneous actions;
$I$ is the total function describing the initial state of the propositions and real variables;
$G$ is the function indicating the goal condition.
\end{definition}

\begin{definition}
A \textbf{durative action} $a$ is a tuple $\langle pre(a), \eff(a), dur(a) \rangle$ where
$pre(a)$ is a set of conditions partitioned into at-start, over-all and at-end conditions;
$\eff(a)$ is the  set  of  action  effects; and
$dur(a)$ is the duration constraint.
An \textbf{instantaneous action} is a tuple $\langle pre(a), \eff(a)\rangle$ where
$pre(a)$ is the set of preconditions; and
$\eff(a)$ is the  set  of  action  effects.
\end{definition}

\noindent A durative (resp. instantaneous) action is \textit{applicable} in a state $S=P\cup V$ if the at-start condition (resp. precondition) of the action is satisfied by $P\cup V$. We also say that the end of a durative action is applicable if the action is currently executing in the state, the action duration constraint is satisfied, and the at-end condition of the action is satisfied by $P\cup V$. Applying an instantaneous action, durative action start, or durative action end $a$ to the state $S$ produces a resultant state $S(a)$.

We formalize plans of actions as networks, similar to Temporal Plan Networks~\cite{kim01}:

\begin{definition}\label{def:plan}
A \textbf{plan} $\Pi$ for a planning problem $\langle P,V,A,I,G \rangle$ is the graph $\langle N,C \rangle$ where each node $n\in N$ represents the plan start, an instantaneous action, or the start or end of a durative action; and each edge $c\in C$ represents a temporal relation: $x < time(n_1) - time(n_2) < y$ for $n_1, n_2 \in N$ and $x,y\in \mathbb{R}$. Each edge $c$ is labelled as either \textbf{causal}, \textbf{interference}, or \textbf{action duration}. Action duration edges express the temporal constraints between the start and end of durative actions. Causal edges express temporal relationships inferred from the causal support between actions. Similarly, interference edges express the temporal relationships inferred from the interference between actions.
\end{definition}

\noindent A plan is \textit{totally-ordered} if there exists only one total ordering of nodes that can satisfy all of the temporal relations. 

Similarly to the definitions presented in \cite{how04} we say that a totally-ordered plan is \textit{executable} if the plan can be simulated by applying each action in order, and all of the prescribed actions are applicable. The plan is \textit{valid} if the final state satisfies the goal condition $G$.

\section{Robust Plan Execution}\label{sec:method}

PDDL2.1 planners output is represented as \textit{time-triggered} plans (e.g.~\cite{col10,ran15}). A time-triggered plan is a set of tuples $\langle t, a, d \rangle$, where $a$ is an action, $t$ is the time at which the action should be executed and $d$ is the prescribed duration.

\begin{itshape}
As a running example, consider a scenario where a fleet of robots can move on a known map and are tasked to retrieve specific items produced from a pool of machines located at known positions. When a robot is at a machine location, it can turn the machine on. Once the machine is on, two robots are required to be at the machine location to produce an item. Clearly, the navigation actions of the two robots are independent until they synchronize to be at a specific machine together.
A concrete instance is as follows. Two robots \texttt{r0}, initially in location \texttt{wp1}, and \texttt{r1}, initially in location \texttt{wp0}, must reach location \texttt{m0} where a machine is located, switch on the machine and produce an item that is finally delivered in location \texttt{wp1}. A valid time-triggered plan $\Pi^{tt}_{ex}$ for the problem is reported below, indicating the starting time for each action to be executed (before the colon) and the expected duration (in brackets)\footnote{This the the usual syntax PDDL planners use for temporal plans.}.
\end{itshape}
\begin{Verbatim}[fontsize=\scriptsize]
   0.000:  (goto r0 wp1 m0)           [14.000]
   0.000:  (goto r1 wp0 m0)           [ 9.000]
   14.001: (switch_on r0 m0)          [ 5.000]
   19.002: (load_at_machine r1 r0 m0) [15.000]
   34.002: (goto r1 m0 wp1)           [14.000]
   48.002: (ask_unload r1 wp1)        [ 5.000]
   53.003: (wait_unload r1 wp1)       [15.000]
\end{Verbatim}

In order to allow for temporal flexibility, time-triggered plans must be converted into partially-ordered plans as per \cref{def:plan}. 
In addition, we further relax the plan into an \emph{adaptable} partially-ordered plan by removing the ordering relations that represent the causal support between actions. This allows for unexpected, but beneficial events in the environment to achieve the preconditions of actions. An adaptable partially-ordered plan can then be executed using the online procedure defined in \cref{sec:com}.

\begin{figure}[tb]
    \centering
    \resizebox{\columnwidth}{!}{\begin{tikzpicture}
  [
    plan/.style = {cylinder, shape border rotate=90, draw=black, minimum height=1.5cm, minimum width=1cm, shape aspect=.25, very thick, align=center, text width=1cm, fill=yellow!40},
    probs/.style = {cylinder, shape border rotate=90, draw=black, minimum height=1.5cm, minimum width=1cm, shape aspect=.25, very thick, align=center, text width=1cm, fill=green!20},
    data/.style = {cylinder, shape border rotate=90, draw=black, minimum width=2.5cm, shape aspect=.25, very thick, align=center, text width=2.5cm},
    tool/.style = {rectangle, draw=black, thick, minimum height=1.5cm, minimum width=2.5cm, very thick,  align=center, text width=2.5cm},
    link/.style = {-latex, very thick}
  ]
        
  \node[data, fill=blue!20] (model) {PDDL 2.1 Planning Problem};
  \node[tool] (planning) [right=1cm of model] {PDDL2.1 Planner};
  \node[plan] (ttplan) [right=1cm of planning] {\Large $\Pi^{tt}$};
  \node[tool] (tostn) [right=1cm of ttplan] {ESTEREL Transformer};
  \node[plan] (plan) [right=1cm of tostn] {\Large $\Pi$};
  \node[tool] (causal) [right=1cm of plan] {Causal Edge Remover};
  \node[plan] (pplan) [right=1cm of causal] {\Large $\Pi'$};
  
  \node[tool] (perc) [below=2cm of model] {Perception};
  \node[tool] (obs) [right=1cm of perc] {State Probability Estimator};
  \node[probs] (s0) [right=1cm of obs] {\Large $S_0$};
  \node[tool] (algo) [right=1cm of s0] {Total-Order Extractor};
  \node[data, fill=red!20, text width=3cm] (tos) [right=1cm of algo] {Totally-Ordered\\Plans};
  \node[tool] (exec) [right=1cm of tos] {Action Dispatcher};

  \draw[link] (model) -- (planning);
  \draw[link] (planning) -- (ttplan);
  \draw[link] (ttplan) -- (tostn);
  \draw[link] (tostn) -- (plan);
  \draw[link] (plan) -- (causal);
  \draw[link] (causal) -- (pplan);
  \draw[link] (pplan) -- ($ (pplan.south) - (0, 0.5) $) -- ($ (algo.north) + (0, 0.5) $) -- (algo);
  \draw[link] (obs) -- (s0);
  \draw[link] (s0) -- (algo);
  \draw[link] (algo) -- (tos);
  \draw[link] (tos) -- (exec);

  \draw[link] (perc) -- (obs);
  
  \draw[dashed, very thick] ($ (model.south west) - (0.5, 1) $) node[anchor=south west]{\large \textsc{Offline}} node[anchor=north west]{\large \textsc{Online}} -- ($ (causal.south east) + (2, -1) $);
  \draw[link, dashed] (exec.south) -- ($ (exec.south) - (0, 1) $) node[anchor=north]{\large Action to execute};
  \draw[link, dashed] ($ (perc.south) - (0, 1) $) node[anchor=north]{\large Sensor data} -- (perc.south);
\end{tikzpicture}}
    \caption{High-level overview of the proposed flow from planning to execution.}
    \label{fig:flow}
\end{figure}
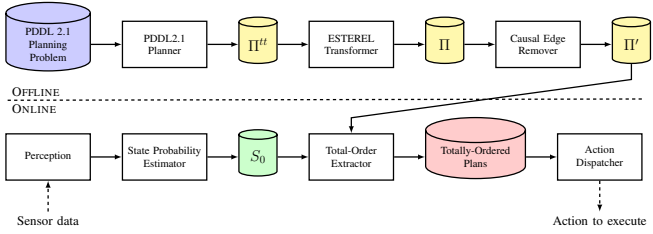

Our approach is essentially composed of three phases, depicted in \cref{fig:flow}. The first phase is performed offline and is responsible for converting a totally-ordered input plan into an adaptable partially-ordered plan (this corresponds to the upper part of \cref{fig:flow} and we provide full details in \cref{sec:gen}).
The adaptable partially-ordered plan is passed to the second, online phase (\cref{sec:com}) that generates the set of valid totally-ordered plans described by the adaptable partially-ordered plan. This phase is indicated as ``Total-Order Extractor'' in \cref{fig:flow}. Finally, the produced set of valid totally-ordered plans is the input for the action dispatcher (\cref{sec:exe}) that chooses the next action to execute by reasoning over the set of valid totally-ordered plans.

\subsection{Phase 1: Generating Adaptable Partially-Ordered Plans}\label{sec:gen}

Once a time-triggered plan ($\Pi^{tt}$ in \cref{fig:flow}) is generated by an off-the-shelf planner, it needs to be converted into an ordinary partially-ordered plan $\Pi$ by generating a node for each instantaneous action and each durative action start and end. The relations are generated as follows.
For each node $n_1$ that supports the condition of a node $n_2$, the temporal relation $0 < time(n_2) - time(n_1) < \infty$ is generated.
For each pair of nodes $\langle n_1, n_2 \rangle$ representing the start and end of a durative action $a$, a relation representing the constraints in $dur(a)$ is generated.
Finally, for each pair of actions in the plan $a_1$ and $a_2$ that interfere, an interference relation is generated.
Two actions interfere if they have conflicting effects, or the effects of one action conflict with the conditions of the other (see \cite{fox03} Definition 12). As durative actions are represented by two nodes (\textit{at-start} and \textit{at-end}), the relation is generated between the nodes that contain the interfering effects or conditions. In the case of an interfering \textit{over-all} condition a relation is generated for both nodes.

\begin{figure}[tb]
    \centering
    \includegraphics[width=.75\columnwidth]{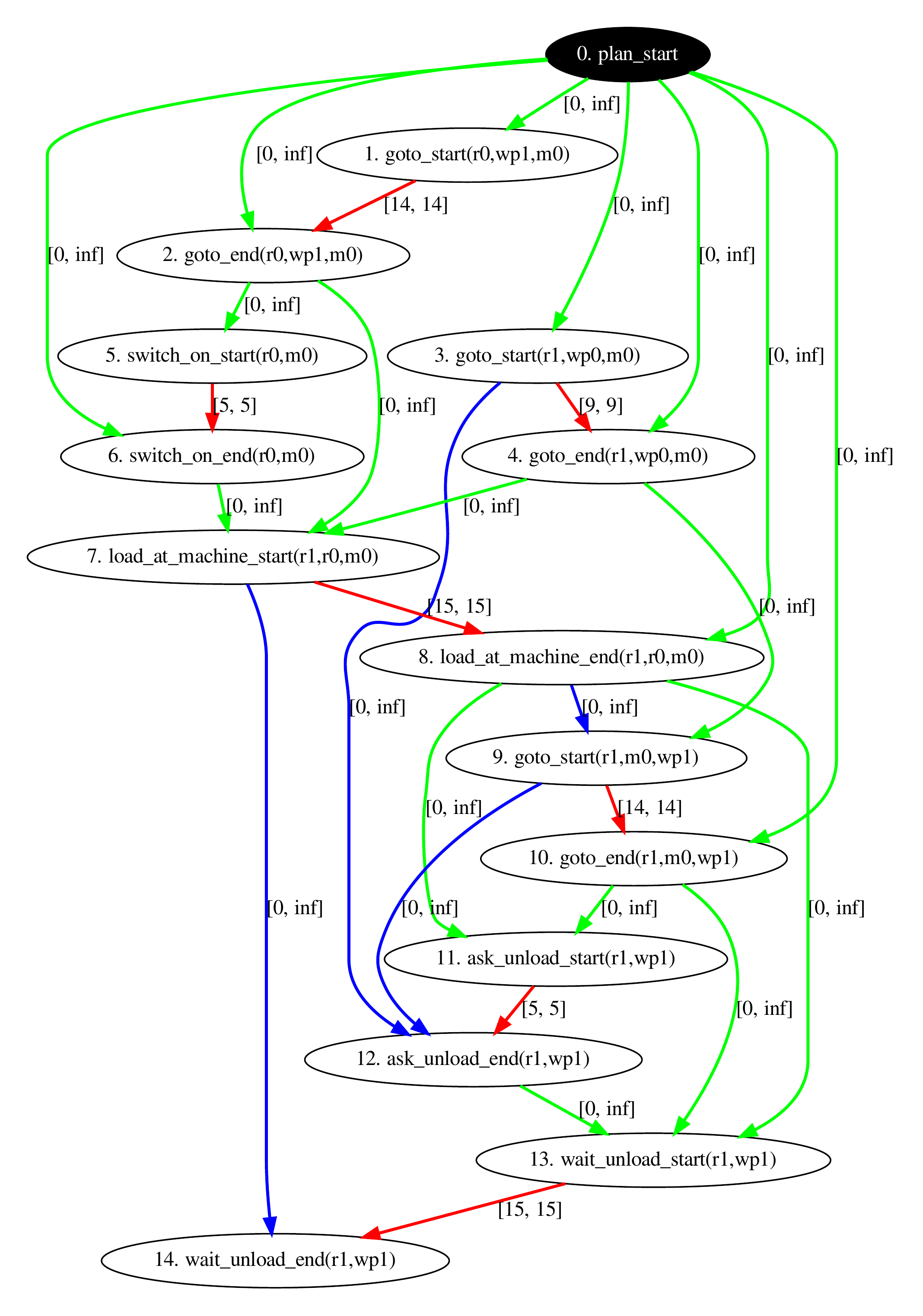}
    \caption{Graph of the de-ordered plan $\Pi_{ex}$. Nodes represent instantaneous actions or startings or endings of durative actions. Edges represent temporal relationships in the form $[min,max]$, and illustrate causal support (green) action duration (red) and interference (blue) constraints.}
    \label{fig:esterel2}
\end{figure}

\begin{itshape}
For our running example, this gives the plan $\Pi_{ex}$, whose graph representation is reported in \cref{fig:esterel2}. It is easy to see that, while multiple ordering of the events are possible, the plan has a well-defined causal structure that ensures a sequence of pre-defined actions.
\end{itshape}

Before execution begins, the plan is relaxed into an adaptable partially-ordered plan. Given a plan $\Pi$ the adaptable partially-ordered plan $\Pi'$ is defined as $\Pi' = \langle N', C' \rangle$, where $N'=N$ and $C'\subseteq C$ contains only the edges of C labelled interference and action duration. By removing the causal support edges, the execution algorithm will be given more flexibility in action selection, possibly skipping actions whose effects have already been achieved by exogenous events. This is achieved using the algorithm in \cref{sec:com}.

\begin{figure}[tb]
    \centering
    \includegraphics[width=\columnwidth]{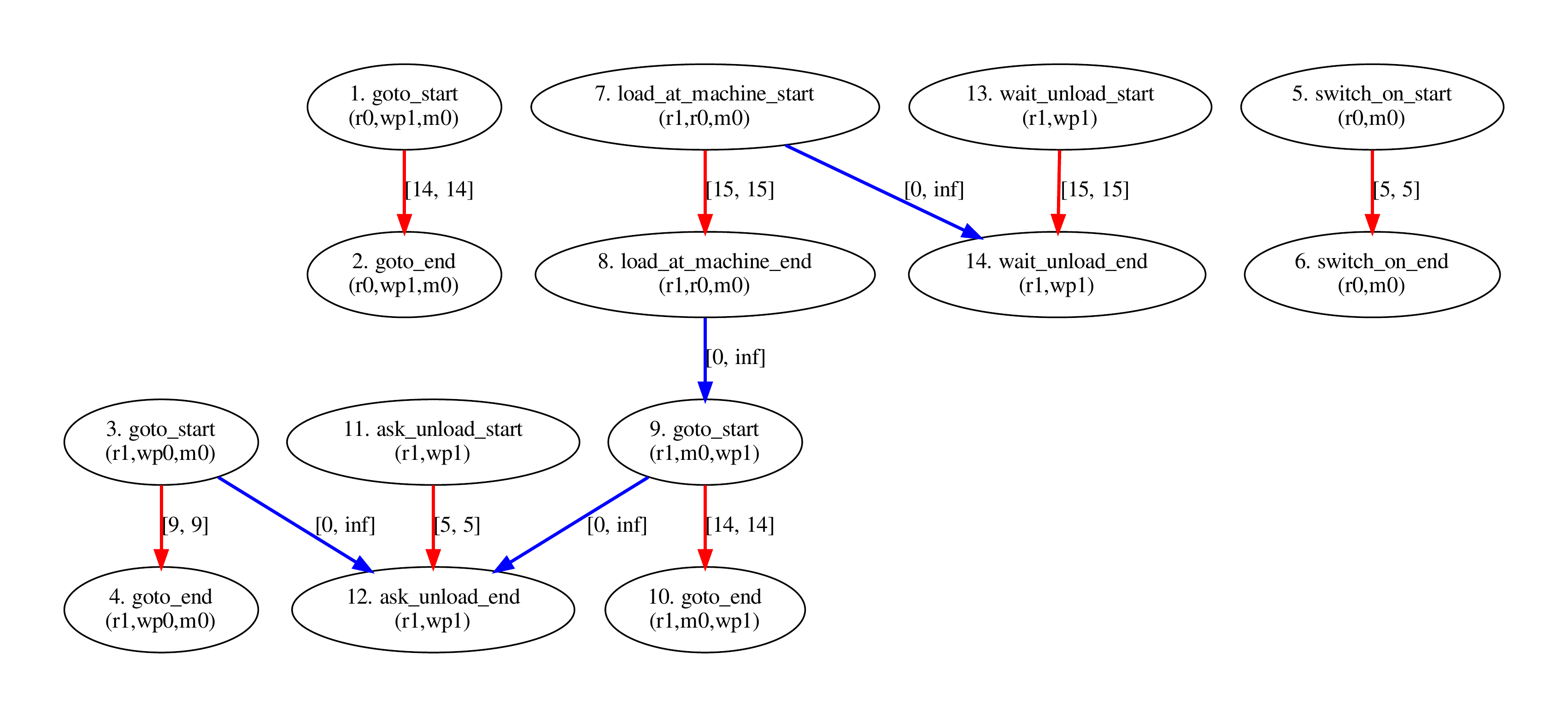}
    \caption{Graph of the adaptable partial-order plan $\Pi'$ obtained from $\Pi$ by removing causal edges.}
    \label{fig:esterel3}
\end{figure}

\begin{itshape}
In our running example, we obtain the adaptable partial order plan $\Pi'_{ex}$ reported in \cref{fig:esterel3}. This graph allows for many more executions orderings with respect to \cref{fig:esterel2}, but causal structure is no longer guaranteed for all of them and we need to restore this structure using runtime reasoning.
\end{itshape}

\subsection{Phase 2: Extracting the Set of Totally-Ordered Plans}\label{sec:com}

Each adaptable partially-ordered plan generated from a valid totally-ordered plan encodes a set of valid totally-ordered plans. Plan execution can be thought of as selecting one such totally-ordered plan to execute. In addition to this, our algorithm considers a notion of uncertainty in the environment, selecting the total-orderings that are most likely to succeed, given a known probability over propositions.


To perform this reasoning, we assign a probability to each predicate in $P$ by means of a map $\prob:P\rightarrow [0;1]$ (this is in fact a fuzzy truth assignment). Then, we consider a set of states of the form $S = \langle P \cup V, \prob\rangle$. Moreover, for each action $a$ we use the map $\psi_a:\eff(a)\rightarrow [0;1]$ to represent the probability of setting a proposition $p\in\eff(a)$ to true after executing the action. For a deterministic action, $\psi_a(p)=1$ if $a$ adds $p$, and $\psi_a(p)=0$ if $a$ deletes $p$.
The operation of applying an action $a$ in a state $S$ (i.e. $S'=S(a)$) is extended to update the set $\prob$ by assigning
$$\prob(p) = \psi_a(p)$$
for all $p\in \eff(a)$, while the $P \cup V$ part of the state is updated as per \cref{sec:met}. As for the transitions, $S'$ can be reached from state $S$ with the joint probability of $a$'s preconditions.

The algorithm requires as input an adaptable partially-ordered plan $\Pi$ and the current state $S_0$ (where we have probabilities assignments in the map $\prob$). The complete set of totally-ordered plans are then generated as described in \cref{algorithm}.
The algorithm can be run before execution, in which case the plan $\Pi$ is the whole adaptable partially-ordered plan and the state $S_0$ is the initial state before starting the execution.
If, instead, the algorithm is run online, $S_0$ is the current state with the probabilities derived from observations and the plan $\Pi$ is the adaptable partially-ordered plan where nodes representing actions that have been executed in the past are removed. In either case, the algorithm will return the set of valid totally-ordered plans that can be executed from the state $S_0$.

\begin{algorithm}[tb]
  \small
  \begin{algorithmic}[1]
 
 \Function{GeneratePlans}{$S_0,\Pi$}
  \State{$O \leftarrow \{n \,| \forall n \in \Pi\}$}
  \Comment{initialize open list.}
  \State{$C \leftarrow \{c \,| \forall c \in \Pi\}$}
  \Comment{set of temporal constraints.}\State{$F \leftarrow \emptyset$}
  \Comment{initialize set of ordered nodes.}
  \State{$R \leftarrow \emptyset$}
  \Comment{return set of totally-ordered plans.}
  \State{\Call{orderNodes}{$O$,$C$,$F$,$S_0$,$R$,$1$}}
  \State{\textbf{return} $R$}
  \EndFunction
  
  \item[]
  
  \Function{OrderNodes}{$O,C,F,S,R,Q$}

  \If{$\neg $ \Call{checkTemporalConstraints}{$F$,$C$}}
    \State{\textbf{return}}
    \Comment{the ordering is not executable.}
  \EndIf

  \If{$S\models G$}
    \State{$R \leftarrow R \cup \{ \langle Q,F \rangle \}$}
    \Comment{add new valid ordering to $R$.}
    \State{\textbf{return}}
  \EndIf

  \State{$\Phi \leftarrow $ \Call{validNodes}{$O$,$S$}}
  \If{$\Phi = \emptyset$}
    \State{\textbf{return}}
    \Comment{the ordering is not valid.}
  \EndIf

  \ForEach{$\langle q, a \rangle \in \Phi$}
    \State{$K \leftarrow \{ b \in O \,|\, b \prec a \}$}
    \Comment{all nodes $b$ ordered before $a$.}
    \State{$F' \leftarrow F \cup \{a\}$}
    \Comment{update current ordering.}
    \State{$O' \leftarrow O \setminus (K\cup \{a\})$}
    \Comment{remove skipped nodes from $0$.}
    \State{$S' \leftarrow S(a)$}
    \Comment{apply $a$ to $S$.}
    \State{\Call{orderNodes}{$O'$, $C$, $F'$, $S'$, $R$, $Q \times q$}}
    \Comment{recurse.}
  \EndFor
  \EndFunction
  
  \end{algorithmic}
  \caption{Generate Totally-Ordered Plans}
  \label{algorithm}
\end{algorithm}

The algorithm describes a search through the nodes contained in the plan. A representation of the state, $S=\langle P\cup V, \prob \rangle$, is used throughout the search to check action conditions and simulate their effects.
At each step of the search, the \textsc{validNodes} procedure is called to return a set of tuples $\langle q, a \rangle$, where $a$ is an applicable next node, and $q$
is the joint probability of that action's preconditions (line 14). The search branches on each element of $\Phi$, applying the action $a$ to $S$ (line 21).
The probability of reaching a node is stored in $Q$ (initially 1), which is $\prod_{a\in F}\prob(pre(a))$ where $\prob(pre(a))$ is the probability of the action's preconditions in the state in which $a$ was applied.
For example, consider an action $a_0$ with precondition $p_0$ and positive effects $p_1,\ldots,p_5$, with $\psi_a(\eff)=1$ for all effects. Action $a_1$ has precondition $p_1\wedge\ldots\wedge p_5$. The probability in the initial state is $\prob(p_i)=0$ for all $i>0$ and $\prob(p_0)=0.5$. $a_0$ is added to $F$ (line 19) and the state is updated (line 21). The effects of $a_0$ set $p_i=1$ in $S'$ for all $i=1\ldots5$. The recursion is then called updating the probability of the next node to $Q=0.5$ (line 22). The joint probability of action $a_1$ from the new search node is $1$. Thus, the probability of being able to apply the sequence $[a_0,a_1]$ from the initial state is $0.5$.

The search has three \textit{backtracking} conditions. 
First, if adding the new node violates a temporal constraint, including both interference and action duration relations, then the current total ordering is discarded (lines 9-10).
Second, if the goal is achieved then the current total ordering is saved (line 12).
Third, if the goal is not achieved and there are no more applicable nodes, then the current total ordering is discarded (lines 15-16).
In the case of any of these three backtracking conditions, the search resets $S$ and tries the next element of $\Phi$ (line 17), so that every \textit{executable} ordering of nodes is explored, and all \textit{valid} orderings are saved.
The returned result is the a of tuples: $\langle Q, \Pi \rangle$ where $\Pi$ is a totally-ordered plan and $Q$ is that plan's probability of success. 


The \textsc{validNodes} procedure forms the expansion step of the search by returning a set of tuples $\langle q, a \rangle$, where $a$ is an applicable next node, and $q$ is the probability that the node is applicable. An action start node is applicable if the \textit{at-start} and \textit{over-all} conditions of $pre(a)$ are true in the state, while an action end node is applicable if its \textit{at-end} conditions are true.
The probability $q$ of the preconditions of node $a$ (indicated as $pre(a)$) in $S$ is computed from $\prob$ as the joint probability of the preconditions of $a$. For example, a node $n$ with precondition $pre(a) = p_1 \wedge \neg p_2$, where $p_1,p_2\in P$ is $q = \prob(p_1)*(1 - \prob(p_2))$.
If $q$ is greater than some threshold (in our implementation this is 0), then the tuple $\langle q, a \rangle$ is added to the return set of \textsc{validNodes}, otherwise it is discarded.

Theoretically, the number of possible total orders induced by an adaptable partially ordered plan is factorial in the number of nodes: it suffices to consider a plan with no constraints where every permutation of the nodes is a valid total order. This is the dominant complexity cost, hence the algorithm runs in $O(n!)$. Nonetheless, this case is practically never encountered for meaningful, practical plans, because interference and duration constraints dynamically prune executions that are causally impossible given the current observations. In \cref{sec:exp} we show that this approach exhibits very good empirical performance.

\subsection{Phase 3: Action Selection Policy}\label{sec:exe}

During execution, the executor is tasked with selecting the next node, whether this is the start of an action, the end of an action, or to await some other external timed event.

Given a set of totally-ordered plans with probabilities, $\langle Q, F \rangle$, our executor takes the ordering with maximum $Q$ and select the node that is first in the order (whether that is to dispatch an action start, wait for an action end).
For example, given the set:
$[\langle 0.5, [a,b,c] \rangle, \langle 0.3, [b,a,c] \rangle, \langle 0.3, [b,c] \rangle]$
The executor would choose node $a$ to be executed first.
Note that in this example there are two orderings beginning with node $b$ both with probability $0.3$, but their probabilities are not necessarily independent. For example, it could be that the probability of being able to apply $a$ and $c$ is always $1$, and that the probability of applying $b$ is $0.5$ after $a$, and $0.3$ otherwise. For this reason, we cannot combine the probabilities of success of the orderings starting with $b$ (that would be $0.6$) and we decide to execute $a$ instead.

\begin{itshape}
Given the plan $\Pi'_{ex}$ for our running example situation, \cref{algorithm} can extract all the possible valid total orderings. Among these, the procedure is clearly able to reconstruct the original, totally-ordered plan $\Pi_{ex}^{tt}$, but other orderings are also possible, depending on the observed probabilities. Let us consider an example situation to clarify the possibilities opened by our approach. Suppose that the machine \texttt{m0} is found to be already switched on (with high probability) upon $r0$ arrival. By running \cref{algorithm}, the orderings having \texttt{(switch\_on r0 m0)} as first action to execute will have very low probability, while the ones having \texttt{(load\_at\_machine r1 r0 m0)} will have high probability. For this reason, our action selection policy, exploiting the result from \cref{algorithm}, will choose to execute \texttt{(load\_at\_machine r1 r0 m0)} and a subsequent successful continuation of the plan will effectively skip the \texttt{(switch\_on r0 m0)} action. This is because the system will reach the goal without ever executing such an action. 
\end{itshape}

\section{Implementation and Evaluation}\label{sec:exp}

\newcommand{\myrt}[1]{\parbox{1cm}{\vskip 15pt \centering #1}}
\newcommand{\myhd}[1]{\multirow{2}{*}{\rotatebox{90}{\bf #1}}}
\newcommand{\myd}[1]{\parbox{1cm}{\vskip 15pt \centering #1}}

\begin{figure*}[tbh]
  \centering
  \begin{subfigure}[b]{.32\textwidth}
    \resizebox{\textwidth}{!}{
        \setlength{\tabcolsep}{2pt}
        \begin{tabular}{rr|cc|cc|cc}
                      &         & \multicolumn{2}{c|}{\bf Coverage} & \multicolumn{2}{c|}{\bf Avg. Replans} & \multicolumn{2}{c}{\bf Avg. \#Actions} \\
                      & \#orders             & 2             & 3            & 2               & 3              & 2                & 3                \\ \hline
\myhd{Deadline-Free}  & \myrt{RO} & \myd{98\%} & \myd{94\%} & \myd{0.9} & \myd{0.9} & \myd{12.5} & \myd{17.2}\\[18pt]
                      & \myrt{RP}    & \myd{98\%} & \myd{97\%} & \myd{1.6} & \myd{2.0} & \myd{13.7} & \myd{19.3}\\[18pt] \hline \hline
\myhd{With Deadlines} & \myrt{RO} & \multicolumn{2}{c|}{\myd{91\%}} & \multicolumn{2}{c|}{\myd{0.7}} & \multicolumn{2}{c}{\myd{17.3}}\\[18pt]
                      & \myrt{RP} & \multicolumn{2}{c|}{\myd{97\%}} & \multicolumn{2}{c|}{\myd{2.9}} & \multicolumn{2}{c}{\myd{19.6-}}\\[18pt]
                                               
        \end{tabular}
    }
    \caption{\label{tab:results}Results for deadline-free tasks (top) and for problems with deadlines (bottom), showing coverage, average number of replans across successful tasks, and average number of actions executed.}
  \end{subfigure}
  \hfill
  \begin{subfigure}[b]{.32\textwidth}
    \includegraphics[width=\textwidth]{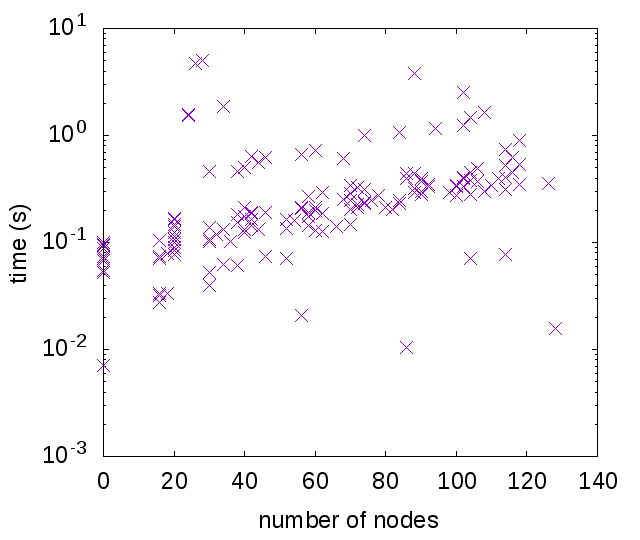}
    \caption{\label{fig:results_nodes}Time taken to produce valid orders for plans of varying size.}
  \end{subfigure}
  \hfill
  \begin{subfigure}[b]{.32\textwidth}
    \includegraphics[width=\textwidth]{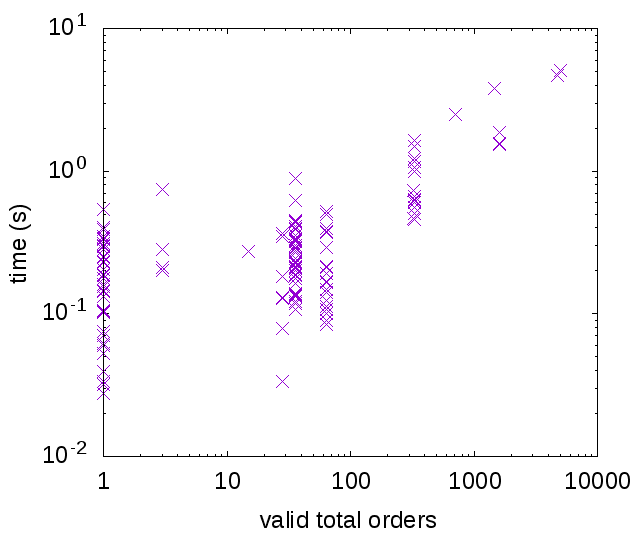}
    \caption{\label{fig:results_plans}The number of valid orders generated, and time taken in generation.}
  \end{subfigure}
  \caption{\label{fig:container} Experimental results.}
\end{figure*}

\begin{figure}[tb]
    \centering
    \includegraphics[width=0.8\columnwidth]{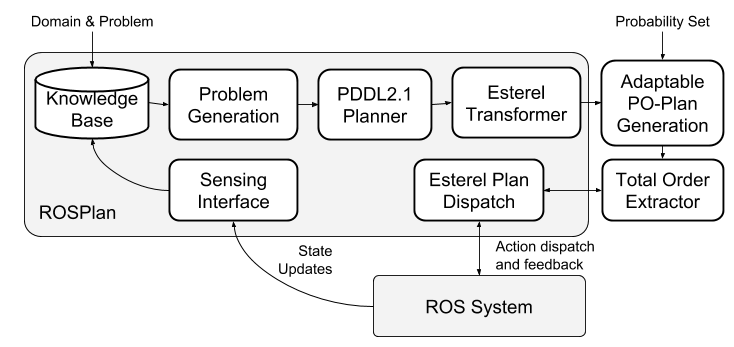}
\caption{System Architecture for the evaluation. The proposed approach has been integrated into the ROSPlan system.}
    \label{fig:arc}
\end{figure}

The approach has been implemented in ROS and integrated as an alternative execution algorithm in ROSPlan~\cite{cas15}. The architecture of the resulting system is illustrated in \cref{fig:arc}. A PDDL2.1 domain and problem file are passed to the system at launch, thereafter a new planning problem is automatically produced during each planning episode.

We use the planner POPF~\cite{col10}, as it is a PDDL2.1 planner that produces time-triggered plans and is already available with ROSPlan. The planner output is parsed into a partially-ordered plan within ROSPlan, and this output is subscribed to by our node, which publishes an adaptable partially-ordered plan following the procedure described in \cref{sec:gen}. This plan is then used to produce a set of valid totally-ordered plans, through an implementation of \cref{algorithm}.

The plan dispatcher selects the first node of the plan with highest probability, and executes that node as described in \cref{sec:exe}. After each node is executed, the totally-ordered plan generation is run again (\cref{algorithm}).

\subsection{Experiment Description}

We use the tasks in the Robot Delivery domain to investigate our approach.
The domain is a much simplified version of the domain used in the Planning and Execution Competition for Logistics Robots in Simulation~\cite{nie15} comprising a fleet of small robots that can navigate in an euclidean graph. These robots are tasked to pick and deliver orders within a deadline. Collecting orders requires two robots present at a machine.
We randomly generated a total of 30 initial states with 3 robots, 3 to 5 machines, 4 to 8 delivery locations, and the goal to deliver 2 or 3 orders. For 9 of these problems, the optimal plan duration was calculated, and a deadline was added to each order equal to $1.5$ times the optimal plan duration. If this deadline is passed, the order cannot be delivered and the task is failed. Thus, the problem set contains 39 tasks overall. 9 tasks with deadlines, and 30 tasks without.

Each task was run in simulation using our re-ordering approach (RO) and a replanning (RP) approach. The RP approach attempts to directly execute the partially ordered plan $\Pi$, and replans when the execution fails. In contrast, RO follows the procedure described above, generating new total orders online to select the next action, and replanning only when no valid total ordering can be found by \cref{algorithm}. Replanning takes place when (1) an action reports failure, (2) an action is to be executed, but its preconditions are not true, and (3) a temporal constraint is violated.

We use a non-physics simulation that includes a probability of action failure and non-deterministic action duration. In addition, the ground truth of the simulation is not static. For example, the proposition \texttt{(machine\_on m0)} may be true in the initial state, but later change due to exogenous events. In this case, the plan execution may fail due to a mismatch between the planner's model and the ground truth, leading to replanning condition (2). Each task and system was run in simulation 10 times to account for random events, action failure, and non-deterministic action duration.

For each run we recorded whether or not the task succeeded in delivering all of the orders. For the tasks that succeeded, we also recorded the number of times the plan execution failed and the system had to replan, and the number of actions that were actually executed in order to achieve the goals.
The results for both the cases are shown in Table~\ref{tab:results}. The results demonstrate our hypothesis that in problems with and without deadlines the re-reordering approach will result in fewer replans, and that fewer actions will be executed overall. The coverage of RP is marginally higher in the problems with deadlines, where the overhead of RO can cause failures.
However, the reduction in the number of replans is significant while impact on coverage is minimal. In domains with dead-ends (from which recovery through replanning is impossible) or in platforms with insufficient computational power to perform online planning, the results show that RO is a viable plan repair.




As discussed in \cref{sec:com}, the worst case complexity of our algorithm can be factorial in the size of the plan. To provide some empirical insight in terms of performance, 160 problems were generated and solved, producing STN plans with up to 128 nodes. The time taken to generate the adaptable partially-ordered plan and to extract all valid total orders was recorded, as well as the number of total orders. \Cref{fig:results_nodes} plots the time against the number of nodes, while \cref{fig:results_plans} shows the time against the number of valid total orders produced. The algorithm takes less than 10 seconds in all cases, hence it is definitely suitable to be used online.



\section{Conclusions}\label{sec:con}

In this paper, we proposed a novel approach for improving the robustness of the execution of automatically-generated task plans with respect to unforeseen circumstances. The approach consists in relaxing the causal structure of the generated plan, allowing for run-time adaptation of the ordering of the actions and, possibly, for the skipping of some action in situations where their execution is no longer needed. The approach is proven effective on a simulated use-case, where the number of re-planning attempts was reduced.

There are several directions for future work. First, we plan to integrate this approach with~\cite{cas19} to natively handle discrepancies in continuous dimensions of the problem, such as time or resources. Second, we would like to try to use our approach with multiple plans: instead of generating the total orders from a single plan, we can use several, diverse plans to allow for more variability in the action selection policy. Finally, we plan to consider other kinds of runtime plan repairs in addition to action reordering and skipping.   

\section*{Acknowledgments}
This work was supported by the FCT projects [UID/EEA/50009/2013] and [PTDC/EEI-SII/4698/2014].

\bibliographystyle{IEEEtran}
\bibliography{bibliography}

\end{document}